\documentclass[10pt,oneside,onecolumn,numbers]{aibuildai}

\usepackage{needspace}

\reporttype{AIBuildAI Research}
\title{AIBuildAI-2.5: Efficient Autonomous AI Model Development Through LLM-Guided Tree Search}
\shorttitle{AIBuildAI-2.5}
\leadauthor{AIBuildAI Team}

\author[1,*]{Peijia Qin}
\author[1,*]{Ruiyi Zhang}
\author[1,*]{Qi Cao}
\author[1,*]{Han Guo}
\author[1]{Li Zhang}
\author[1,2,\Letter]{Pengtao Xie}

\affiliation[1]{Department of Electrical and Computer Engineering, University of California San Diego, La Jolla, CA 92093, USA}
\affiliation[2]{Department of Medicine, University of California San Diego, La Jolla, CA 92093, USA}

\contribution[*]{Equal contribution.}
\contribution[\Letter]{Corresponding author.}
\correspondence{\email{p1xie@ucsd.edu}}
\codeurl{https://github.com/aibuildai/AI-Build-AI}

\abstract{Autonomous agents that automatically build artificial intelligence (AI) models could broaden access to AI across science and engineering. A popular line of such agents frames model building as a code search problem and solves it by tree search, in which each node is a candidate program and the tree grows by generating a child program from a parent, and these agents now approach the capability of experienced AI engineers on realistic benchmarks. However, these agents have weaknesses in efficiency, which are especially consequential given the resource-heavy nature of the task: training candidate models demands substantial computation and time, and the many LLM calls incur a high inference cost. Three such weaknesses have not been fully addressed by existing agents. First, only a small number of candidates can be executed within a realistic budget, so search rules that rank nodes by executed rewards, such as Monte Carlo-style tree search, rely on few and noisy scores and select the next node to explore less effectively. Second, no resource-aware strategy is used to schedule training jobs, which can lower hardware utilization and training efficiency. Third, every agent call is served by a single powerful model, which inflates inference cost. Here we introduce AIBuildAI-2.5, an agentic system that carries out the tree search with LLM agents and addresses each of the three issues. AIBuildAI-2.5 proposes a novel LLM-guided tree search, in which LLM agents provide a prior estimate of which search directions are promising, both locally and globally: a judge scores each waiting candidate on several aspects, namely its expected improvement, its grounding in the measured results of its parent, and its feasibility, and a selector ranks the entire pool of candidates from these scores and the state of the search. This prior, drawn from the knowledge of the LLM and combined with the rewards measured so far, allows the most promising candidates to be executed first. In addition, AIBuildAI-2.5 comprises a scheduler that launches training jobs with the current hardware resource status taken into account and a router that assigns lower-cost LLMs to less demanding tasks while reserving the most capable LLM for the most challenging sub-tasks in the AI model building workflow. AIBuildAI-2.5 ranks first on MLE-Bench with a medal rate of 73.3\%, and outperforms a strong autonomous baseline on all six autonomous AI research tasks from AIRS-Bench on which it is evaluated. Moreover, model routing matches or exceeds the task scores of a configuration that runs every role on the most capable model while incurring less than half of its cost. Together, these results demonstrate that LLM-guided, resource-aware tree search and cost-aware model routing enable efficient, expert-level autonomous AI model development.}

\begin{document}
\maketitle

\section{Introduction}

Building a competitive AI model for a given task is costly and reliant on specialized expertise, motivating a growing line of autonomous agents based on large language models (LLMs) that, given only a task description and a dataset, build such a model end-to-end~\cite{toledo2025airesearchagentsmachine,jiang2025aide,yang2025rdagent,zhang2026aibuildai}. These agents have advanced rapidly on realistic AI development benchmarks such as MLE-Bench~\cite{chan2025mlebench}, increasingly approaching the capability of experienced AI engineers on a broad range of Kaggle-style~\cite{kaggle} tasks. Among them, a popular line of methods frames automated AI model building as a code search problem, in which the agent searches a space of candidate programs, each a training-and-inference pipeline, for the one whose trained model scores best on a validation set. Several search algorithms have been applied to this space, including evolutionary search~\cite{zhu2026mlmaster2}, parallel multi-start search~\cite{zhang2026aibuildai}, and tree search~\cite{jiang2025aide,du2026mlevolve}, of which tree search has become the most widely adopted. A tree search maintains a tree in which each node is a candidate program and grows it by generating a child program from a parent, for example by improving a promising node or debugging a broken one~\cite{du2026mlevolve}, and then selecting which node to expand next. Tree search is appealing because every new candidate is derived from an earlier one and thus inherits what previous attempts have learned, while separate branches explore different directions independently of one another. Tree search methods of this kind have achieved strong performance on automated AI model building.

Despite their effectiveness, these agents share a limitation in efficiency. Building an AI model is an exceptionally resource-heavy process: producing and assessing a single candidate requires implementing a full training and inference pipeline and then training and validating it, which consumes hours of wall-clock time with heavy GPU and CPU usage, together with the inference cost of a large number of LLM calls. This resource-intensive nature gives rise to three issues that previous methods have not fully addressed. First, only a small number of candidates can be executed within a realistic budget, which leaves the search with few and noisy reward scores. Current agents select nodes with rules borrowed from classical search, most prominently Monte Carlo-style tree search~\cite{du2026mlevolve} and greedy best-first heuristics~\cite{jiang2025aide}, which guide the search by the rewards of executed nodes, and with such a sparse signal these rules select the next node to explore less effectively. Second, they lack a resource-aware strategy for scheduling training jobs, which can lower hardware utilization and training efficiency. Third, LLM inference is used inefficiently: every agent call, from designing and coding a solution to evaluating candidates, is served by a single powerful language model, even though many calls are routine and could be served by a far cheaper model, which inflates cost without improving the result.

To address these limitations, we introduce AIBuildAI-2.5, an agentic system for efficient autonomous AI model development. AIBuildAI-2.5 formulates AI model building as a code search problem and solves it by tree search, in which each node is a candidate program and the tree grows by expanding a node into child programs. Node expansion is carried out by LLM agents: a designer proposes the initial candidate programs, a reviser proposes revisions of a finished program, and a coder implements each proposal as a new node. The main contribution of AIBuildAI-2.5 addresses the first limitation with a stronger selection strategy. Instead of a rule that ranks nodes by executed rewards alone, an LLM judge scores each waiting candidate on three evidence-grounded dimensions, namely its expected improvement over its parent, how well its rationale is anchored in the measured results of the parent, and its feasibility, and an LLM selector then ranks the whole pool from these scores and the state of the search, so that the most promising candidates are executed first. These decisions give each candidate a strong prior estimate drawn from the knowledge of the LLM and combine it with the rewards measured so far, which steers the search toward stronger solutions while reserving the scarce compute for the candidates most likely to benefit from it. To address the second limitation, a scheduler launches training jobs while taking the current state of the hardware resources into account. To address the third limitation, a router assigns each agent role the least costly model that can still perform it well, reserving the most capable model for the most challenging roles, and grounds these choices in a router knowledge system that records which models have sufficed for which roles in past runs.

We evaluate AIBuildAI-2.5 across standardized benchmarks and autonomous AI research tasks. On MLE-Bench~\cite{chan2025mlebench}, a benchmark of 75 realistic Kaggle-style AI development tasks spanning visual, textual, time-series, and tabular modalities, AIBuildAI-2.5 ranks first on the leaderboard~\cite{mlebench_commit_2026} with a medal rate of 73.3\%, surpassing all existing autonomous AI development systems, including MARS~\cite{chen2026mars}, ML-Master~\cite{liu2025mlmaster}, InternAgent~\cite{team2025internagent}, R\&D-Agent~\cite{yang2025rdagent}, AIDE~\cite{jiang2025aide}, AIRA-dojo~\cite{toledo2025airesearchagentsmachine}, MLEvolve~\cite{du2026mlevolve}, and AIBuildAI~\cite{zhang2026aibuildai}. On AIRS-Bench~\cite{lupidi2026airsbench}, a suite of autonomous AI research tasks in which the agent must build a modeling pipeline from a problem specification alone, with no baseline code, AIBuildAI-2.5 attains a better score than MLEvolve~\cite{du2026mlevolve} on all six tasks we evaluate, spanning molecular, temporal, and linguistic modalities. In addition, per-role model routing matches or exceeds the task scores of a configuration that runs every agent role on the most capable model, at less than half of its cost. Together, these results show that pairing LLM-guided, resource-aware tree search with cost-aware model routing enables efficient autonomous AI model building that derives competitive models while reducing computation and LLM inference cost.

\begin{figure*}[!htb]
\centering
\aibuildgraphic[width=0.9\linewidth]{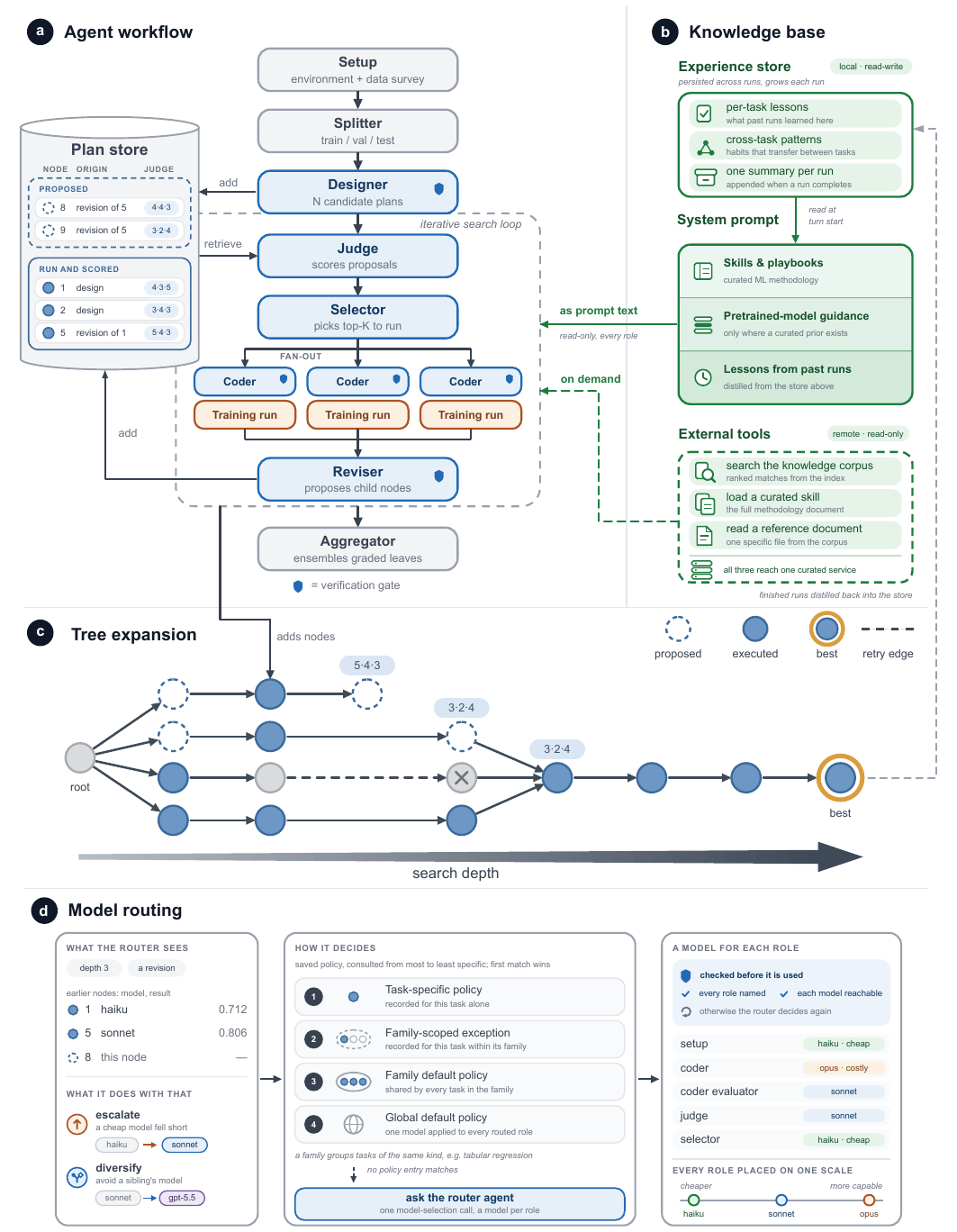}
\caption{\textbf{Overview of AIBuildAI-2.5.}
\textbf{a}, Agents. A designer proposes initial plans, a coder implements each proposal as a node and debugs it until it runs, and a reviser reads a finished node and proposes revisions. Auxiliary agents include a setup agent, a data-split agent, and an aggregator.
\textbf{b}, Knowledge base. Agents draw on a hierarchical knowledge base of AI modeling skills and playbooks, and use external tools to retrieve papers and reference documents.
\textbf{c}, LLM-guided tree search. A node is expanded by the reviser proposing revisions and the coder implementing them. A node is selected by a judge that scores each candidate on expected improvement, grounding, and feasibility, and a selector that ranks the pool from these scores and the state of the search.
\textbf{d}, Model routing. A router assigns lower-cost LLMs to less demanding agent roles and the most capable LLM to the most challenging ones, guided by a router knowledge system of past routing outcomes.
}
\label{fig:AIBuildAI_workflow}
\end{figure*}

\Needspace*{6\baselineskip}
\section{Results}
\subsection{Overview of AIBuildAI-2.5}

AIBuildAI-2.5 takes as input a natural-language description of an AI task together with its dataset, and outputs trained model checkpoints and an inference script that generates predictions on unseen test data (Fig.~\ref{fig:AIBuildAI_workflow}a). It formulates this task as a code search problem: among a space of candidate programs, each a training-and-inference pipeline, find the one whose trained model attains the highest score on a held-out validation set. AIBuildAI-2.5 solves this problem with multiple LLM agents, each with a distinct role and a distinct output artifact. The designer reads the task and proposes a batch of deliberately varied modeling strategies. The coder turns one strategy, or one revision instruction, into code, writing the training-and-inference program and debugging it until the program runs end-to-end. The reviser reads a completed candidate, including its code, score, and logs, and proposes the next change, either one or more improvements when the candidate succeeded or a single targeted fix when it failed. Three auxiliary agents complete the workflow: a setup agent that prepares the software environment, a data-split agent that creates the validation set on which every candidate is scored, and an aggregator that ensembles the strongest candidates into the final model at termination. Throughout the run, the agents draw on a knowledge base (Fig.~\ref{fig:AIBuildAI_workflow}b), following the knowledge system introduced in AIBuildAI-2~\cite{zhang2026aibuildai2}. The knowledge base provides two types of source: a hierarchical collection of AI modeling knowledge, such as skills and playbooks that codify modeling strategy, and external tools that retrieve papers and reference documents from a curated corpus on demand.

AIBuildAI-2.5 conducts the search as a tree search (Fig.~\ref{fig:AIBuildAI_workflow}c). Each node of the tree is a candidate program, and the tree grows by generating child programs from a parent, so that every new candidate inherits what previous attempts have learned while separate branches explore different directions. Within this framework, the search consists of two operations: expanding a node into child programs, and selecting which node to expand next. For expansion, the designer seeds the tree with its initial strategies, which the coder implements as the children of the root, and thereafter the reviser reads a finished node and proposes revisions, each of which the coder implements as a new child. Each implemented node is then executed to train its model, and its score on the validation set is recorded as its reward. Once a node has received its reward, it can be passed to the reviser in turn, and repeating this loop keeps expanding the tree. For selection, the search maintains a pool of candidate nodes, namely the leaves of the current tree that have not yet been executed, and two agents rank this list so that the most promising leaf is executed first. The judge scores each proposed candidate on three dimensions, namely its expected improvement over the parent, how well its rationale is anchored in the measured results of the parent, and its feasibility within a single training slot. Each score is grounded in evidence already present in the tree, such as the outcomes of similar changes on sibling nodes, and any candidate that exhibits reward hacking behavior receives a score of zero. The selector then reasons over the entire pool, ranking the candidates to run next from the judge scores, the current tree, and the remaining resources. In doing so, it favors architectural diversity, discounts lineages whose parents have reached dead ends, avoids near-duplicates, and shifts from exploration to exploitation as the budget decreases. Finally, a scheduler launches the ranked jobs while taking the current hardware resource status into account, so that several candidates train concurrently without exceeding the available compute.

In addition, AIBuildAI-2.5 comprises a model routing system that reduces the LLM inference cost of the search (Fig.~\ref{fig:AIBuildAI_workflow}d). The workflow is made up of many agent calls, and how demanding each call is varies with both the task and the agent role: one task may hinge on implementing and debugging a long training pipeline, so that the coder needs the strongest model, whereas another has a routine pipeline and hinges on deciding what to try next, so that the designer and reviser do instead. Serving every call with a single frontier model ignores this variation and overpays for most calls. AIBuildAI-2.5 therefore employs a router agent that, given the task, the data, and the current state of the search, assigns each agent role the cheapest model from a pool ranging from lightweight to frontier that it judges sufficient to reach the best outcome the task admits, and adjusts these choices within a run according to how earlier nodes coded by each model have fared. These judgments are further grounded in a router knowledge system that accumulates across runs which models have proven sufficient for which roles under which task conditions, so that routing decisions improve as more tasks are completed.

\begin{figure*}[!t]
\centering
\aibuildgraphic[width=.8\linewidth]{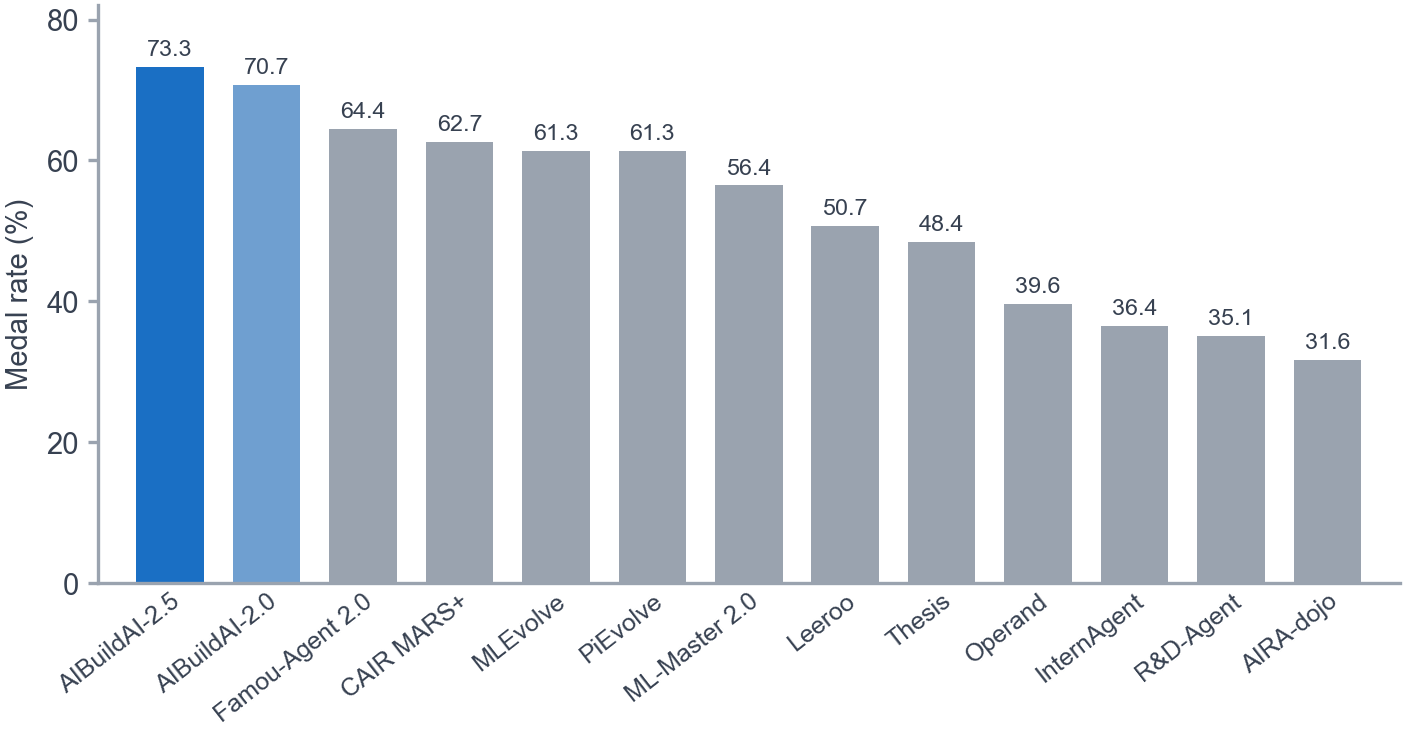}
\caption{\textbf{AIBuildAI-2.5 ranks first on the MLE-Bench leaderboard.} Overall medal rate of AIBuildAI-2.5 compared with recent baseline methods (or method variants) on MLE-Bench. AIBuildAI-2.5 achieves the top overall position with a medal rate of 73.3\%, surpassing prior systems including MARS, Famou-Agent, ML-Master, Leeroo, InternAgent, R\&D-Agent, AIRA-dojo, MLEvolve, and AIBuildAI.}
\label{fig:mle-bench}
\end{figure*}

\subsection{AIBuildAI-2.5 ranks first on the MLE-Bench leaderboard}

We evaluate AIBuildAI-2.5 on MLE-Bench~\cite{chan2025mlebench}, one of the most comprehensive benchmarks for autonomous AI systems on realistic Kaggle-style tasks that require end-to-end AI model development. MLE-Bench comprises 75 tasks curated from past Kaggle competitions, each providing raw datasets, evaluation metrics, and submission protocols faithful to real-world AI development, and spanning visual, textual, time-series, and tabular modalities. Performance is measured using the medal system derived from human Kaggle rankings, in which a submission earns a medal when it meets predefined percentile thresholds on the competition leaderboard. Because these competitions draw a global community of expert practitioners, the medal rate serves as a direct proxy for AI development capability on par with top human practitioners. AIBuildAI-2.5 ranks first on the MLE-Bench leaderboard~\cite{mlebench_commit_2026} with a medal rate of 73.3\% (Fig.~\ref{fig:mle-bench}), establishing a new state of the art and outperforming all recent autonomous AI development systems, including MARS~\cite{chen2026mars}, Famou-Agent~\cite{li2025fmagent}, ML-Master~\cite{liu2025mlmaster}, Leeroo~\cite{nadaf2026kapso}, InternAgent~\cite{team2025internagent}, R\&D-Agent~\cite{yang2025rdagent}, AIRA-dojo~\cite{toledo2025airesearchagentsmachine}, MLEvolve~\cite{du2026mlevolve}, and AIBuildAI~\cite{zhang2026aibuildai}. This top ranking, attained without any task-specific customization, demonstrates that the efficiency of AIBuildAI-2.5 translates into models competitive with top human practitioners across a broad range of real-world AI development tasks.

The leaderboard further demonstrates the efficiency advantage of AIBuildAI-2.5. AIBuildAI-2.5 surpasses MLEvolve~\cite{du2026mlevolve}, the other tree-search system on the leaderboard, which selects the next node to run with Monte Carlo-style statistics. Because MLE-Bench fixes the per-task budget across systems, this lead reflects the effectiveness of LLM-guided tree search and resource-aware scheduling under a fixed constraint on time and computation. The former explores more promising nodes and the latter completes more training runs along those promising directions, which together contribute to the improved overall performance.

\subsection{AIBuildAI-2.5 generalizes to autonomous AI research tasks}

We next evaluate whether AIBuildAI-2.5 extends to autonomous AI research tasks, beyond the general AI development tasks of MLE-Bench. We evaluate on six tasks from AIRS-Bench, a benchmark built to quantify the autonomous research ability of LLM agents in machine learning~\cite{lupidi2026airsbench}. Each task is derived from a machine learning paper and specifies a problem, a dataset, and a metric together with the state-of-the-art value reported in that paper, and the agent receives only this specification and the data, with no baseline code. The six tasks span three categories, with two tasks in each: predicting the heat capacity at constant volume ($C_v$) of a molecule on QM9~\cite{ramakrishnan2014qm9} and its constrained solubility on ZINC~\cite{sterling2015zinc} in the Molecules and Proteins ML category, forecasting weekly photovoltaic power generation and daily Wikipedia page traffic from the Monash repository~\cite{godahewa2021monash} in the Time Series category, and recognizing entailment between sentence pairs on SICK~\cite{marelli2014sick} (SICK-NLI) and predicting the star rating of a Yelp review~\cite{zhang2015character} in the Text Classification category. The published references for these tasks are equivariant and graph-attention architectures for molecules, time-series foundation models, and fine-tuned pretrained language models, respectively.

\begin{figure*}[!t]
\centering
\aibuildgraphic[width=\linewidth]{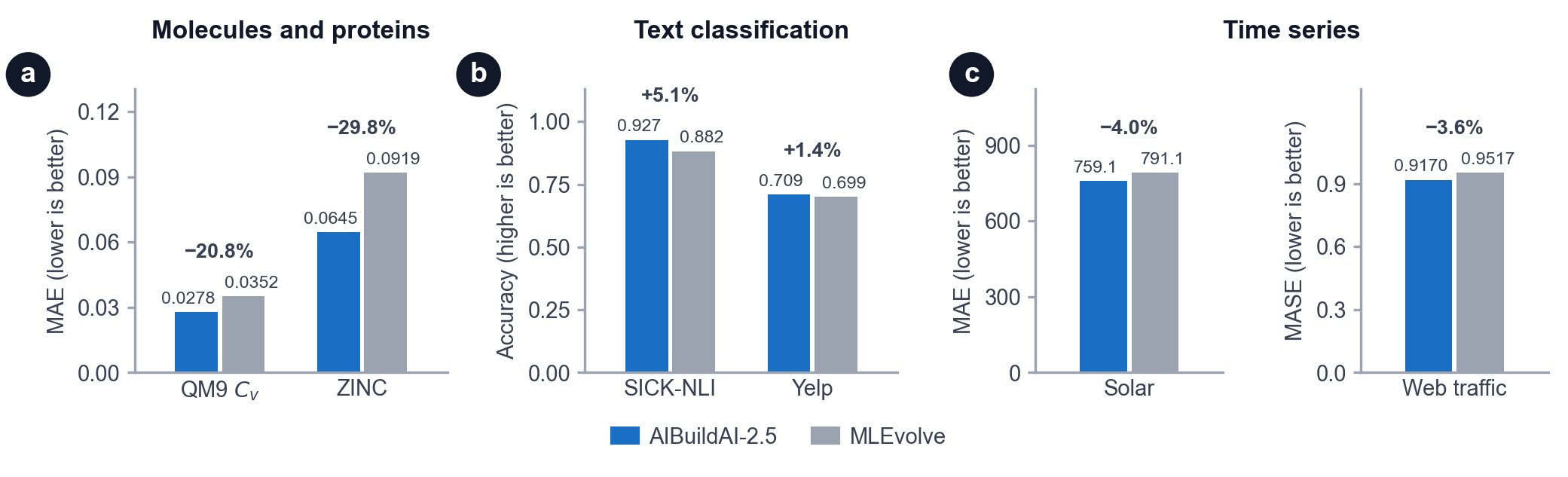}
\caption{\textbf{AIBuildAI-2.5 outperforms MLEvolve on six AIRS-Bench tasks.} Test scores of AIBuildAI-2.5 and MLEvolve on six AIRS-Bench tasks, with the relative margin of AIBuildAI-2.5 annotated above each pair. Panels are grouped by AIRS-Bench task category.
\textbf{a,} Molecules and Proteins ML: heat capacity prediction on QM9 (QM9-$C_v$) and constrained solubility prediction on ZINC, both scored by mean absolute error (MAE; lower is better).
\textbf{b,} Text Classification: natural language inference on SICK (SICK-NLI) and review rating prediction on Yelp, both scored by accuracy (higher is better).
\textbf{c,} Time Series: photovoltaic power forecasting (Solar), scored by MAE, and Wikipedia page traffic forecasting (Web traffic), scored by mean absolute scaled error (MASE). Lower is better for both.}
\label{fig:airs-mlevolve}
\end{figure*}

AIBuildAI-2.5 attains the better score on all six tasks than MLEvolve~\cite{du2026mlevolve}, a competitive autonomous AI model building agent based on heuristic tree search (Fig.~\ref{fig:airs-mlevolve}). It lowers MAE by $20.8\%$ on QM9-$C_v$ and by $29.8\%$ on ZINC (Fig.~\ref{fig:airs-mlevolve}a), improves accuracy by $5.1\%$ on SICK-NLI and by $1.4\%$ on Yelp (Fig.~\ref{fig:airs-mlevolve}b), and lowers solar power MAE by $4.0\%$ and web traffic MASE by $3.6\%$ (Fig.~\ref{fig:airs-mlevolve}c). AIBuildAI-2.5 thus significantly outperforms the MLEvolve baseline on all six tasks, showing that it generalizes from the general AI development tasks of MLE-Bench to autonomous AI research tasks and indicating its potential for AI research. These results further demonstrate the advantage of the AIBuildAI-2.5 strategy of LLM-guided tree search and resource-aware scheduling. The two agents were evaluated under controlled settings with the same budgets and hardware, so that the difference in scores is attributable to the search strategy itself, which clearly shows the effectiveness of AIBuildAI-2.5.

\subsection{Model routing matches frontier-model performance at a fraction of the cost}

\begin{figure*}[!t]
\centering
\aibuildgraphic[width=\linewidth]{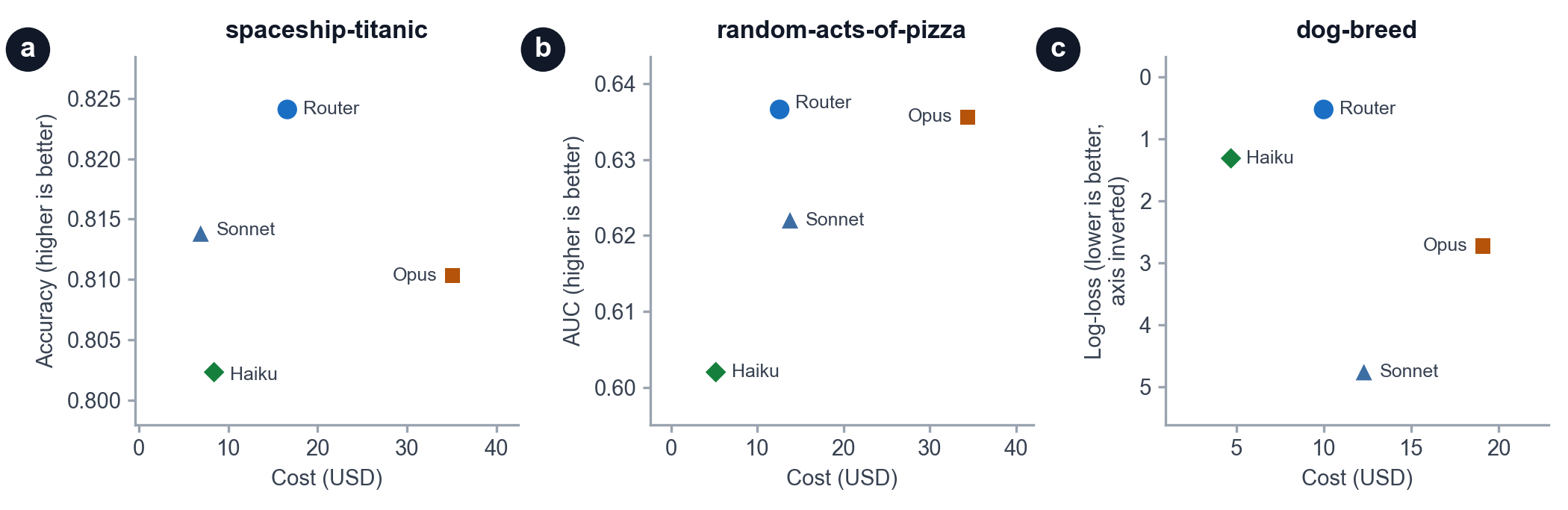}
\caption{\textbf{Model routing combines strong task scores with lower cost.} Cost against task score on three MLE-Bench tasks, one panel per task: \textbf{a,} spaceship-titanic (accuracy); \textbf{b,} random-acts-of-pizza (AUC); \textbf{c,} dog-breed (log-loss). Each point is one AIBuildAI-2.5 configuration, placed at the total cost it incurred in USD (horizontal) and the score it attained (vertical): AIBuildAI-2.5 with model routing (\emph{Router}) and three fixed-model baselines that run every role on a single model of one cost tier, a high-cost model (\emph{Opus}), a medium-cost model (\emph{Sonnet}), and a low-cost model (\emph{Haiku}). In panel \textbf{c} the metric is lower-is-better and the vertical axis is inverted, so that in every panel a better configuration sits higher and a cheaper one further left, and the most desirable position is the upper left.}
\label{fig:router}
\end{figure*}

We next examine whether the AIBuildAI-2.5 routing system achieves strong task scores while reducing the cost incurred by serving every call with a single expensive model. The model pool spans three cost tiers, represented by a high-cost model (Claude Opus), a medium-cost model (Claude Sonnet), and a low-cost model (Claude Haiku). We compare the full AIBuildAI-2.5 router against three fixed-model baselines, each running every agent role on a single tier, on three MLE-Bench tasks that between them cover tabular, textual, and image data: spaceship-titanic, random-acts-of-pizza, and dog-breed. For each configuration we record the score under the task metric and the total cost of the run, and plot the two against each other (Fig.~\ref{fig:router}). All costs are reported in US dollars (USD).

Routing attains scores comparable to or better than the most competitive fixed tier on all three tasks at a lower cost than Opus on each, and therefore lies in the upper-left region of every panel of Fig.~\ref{fig:router}. On spaceship-titanic, routing attains accuracy comparable to the most competitive tier at less than half of the cost of Opus, and earns a gold medal and the top leaderboard rank. On random-acts-of-pizza, it attains an AUC comparable to that of Opus at about one third of the cost. On dog-breed, routing attains the lowest log-loss among all configurations, 60\% lower than that of Haiku, the most competitive fixed tier on that task, at about half of the cost of Opus. Summed over the three tasks, routing reduces cost by 56\% relative to Opus while remaining comparable to or better than the most competitive tier on every task. Haiku runs more cheaply than routing on every task, and Sonnet on spaceship-titanic, but both sacrifice performance, most visibly Haiku on random-acts-of-pizza and Sonnet on dog-breed.

These results demonstrate the effectiveness of the routing system in AIBuildAI-2.5. Routing attains scores comparable to or better than the most competitive fixed tier on every task while incurring less than half of the cost of the frontier-model configuration, whereas the cheaper fixed tiers cost less but lose accuracy. Two mechanisms of the router account for this result. First, the router assigns each agent role the cheapest model it judges sufficient from the task, the data, and the state of the search, grounded in the router knowledge system, so that lower-cost models serve less demanding roles while the most capable model is retained for the most challenging ones. The 56\% cost reduction relative to Opus reflects this assignment. Second, the assignment is specialized per node within a run, escalating the coder to a stronger model when a node revises an ancestor coded poorly by a cheaper model and diversifying away from the model a sibling used. The dog-breed result, where routing attains a lower log-loss than any fixed tier, shows the effect of this adaptation, as a higher-cost model, which spends more reasoning effort, may consume unnecessary time on easy sub-tasks that counts against the total budget of the run.

\section{Discussion}

Across both a comprehensive standardized benchmark and a suite of autonomous AI research tasks, AIBuildAI-2.5 attains a level of autonomous AI model development that matches or exceeds strong prior systems: it ranks first on MLE-Bench with a medal rate of 73.3\%, and on AIRS-Bench it outperforms a competing autonomous agent on all six tasks. In addition, model routing matches or exceeds the scores of an all-frontier configuration at less than half of its cost. These gains come from the novel resource-aware mechanisms of AIBuildAI-2.5, which increase the efficiency of the search and reduce its cost. Since each node reward requires a costly training run, only a small number of rewards can be obtained within the budget. Search rules adapted from classical search, such as Monte Carlo-style tree search~\cite{du2026mlevolve} and greedy best-first heuristics~\cite{jiang2025aide}, were conceived for regimes with orders of magnitude more evaluations, where visit counts and backed-up rewards become statistically meaningful, and at this scale reducing each candidate to a single scalar reward leaves too sparse a signal to distinguish good candidates from poor ones. AIBuildAI-2.5 therefore relies on LLM-guided tree search, in which a judge and a selector provide a prior estimate of the promise of each candidate from its expected gain, its grounding in measured results, and its feasibility, together with the full tree, the results measured so far, and the resources that remain. This estimate directs the scarce evaluations toward the candidates most likely to improve the solution. Resource-aware scheduling then completes more of these evaluations within the same time and computation, and model routing lowers the LLM inference cost of every agent call, freeing budget for additional exploration. Together, these mechanisms enable AIBuildAI-2.5 to obtain more and better solutions from the same computation and inference budget than agents that optimize neither.

A natural direction for further improving AIBuildAI-2.5 lies in its resource scheduler, which uses heuristic rules to decide whether to launch the next job according to the current resource status. Three extensions are promising. First, a growing body of systems research studies more sophisticated strategies for sharing a machine among deep-learning training jobs, including collocating several jobs on one GPU through multiprocessing, NVIDIA's Multi-Process Service, or Multi-Instance GPU partitioning~\cite{robroek2024collocation}, interference-aware collocation that predicts which jobs can profitably share a device~\cite{luo2026castor}, and fine-grained sharing primitives such as job switching, packing, and migration~\cite{xiao2018gandiva,yu2020salus}. Such strategies could raise hardware utilization when individual jobs do not saturate a device. Second, a profiler could estimate the resource consumption of a job before it is executed. Such an estimate would first allow the scheduler to decide more accurately whether a job can be launched under the current resource status. It would further allow the scheduler to adjust the launch order, for example by letting a lighter job run first when the highest-ranked job requires more resources than are currently free. Third, the generated jobs themselves are largely unaware of resources and hardware, since only the scheduler accounts for them. Making the designer, coder, and reviser resource-aware could further improve efficiency, so that the programs they produce are adapted to the available hardware. One way to reach this is a knowledge base of hardware and resource optimization for these agents, from which they can draw efficient methods and implementations.

Beyond building AI models, the framework of AIBuildAI-2.5 applies directly to autonomous research code generation, in which LLM agents write and iteratively refine the code of scientific experiments themselves. AlphaEvolve and FunSearch evolve programs that solve open mathematical and algorithmic problems~\cite{Novikov2025AlphaEvolveAC,FunSearch2024}, and lightweight harnesses such as autoresearch let an agent improve a model-training script unattended over many trials~\cite{karpathy2026autoresearch}. These systems share the property that motivates our design: every candidate must be executed, often by training a model or running a simulation, which consumes resources and time. Each of them can be cast as a code search problem, with an experiment script as the node and its measured outcome as the reward, and AIBuildAI-2.5 can therefore be applied to them with little change. Its advantage in this setting is that it takes these resource constraints into account and explores the space of candidates more efficiently through LLM-guided tree search. Moreover, since scientific discovery often involves long-running experiments, routing agent calls to lower-cost models reduces the inference cost accumulated over a run, so that such agents stay efficient even as the experiments they run grow more expensive.

More broadly, many problems in scientific discovery can potentially be formulated as code search, in which candidate solutions are expressed as programs and scored by an automatic evaluator~\cite{wang2023scientific}. Examples include the discovery of algorithms and numerical methods, the optimization of mathematical constructions, the computational screening of molecules, materials, and proteins, the construction of simulations in physics and climate science, and the design of data-analysis pipelines in biology and medicine~\cite{tbscience2026,wang2026naturebench}. In each of these areas, progress increasingly depends on searching a large space of candidate programs under a tight computational budget, since evaluating a candidate may require a long simulation or an expensive computation. The general idea of AIBuildAI-2.5 is well suited to this setting: LLM agents provide a prior estimate of which directions are promising before the cost of evaluation is paid, resource-aware scheduling extracts more evaluations from the available hardware, and model routing keeps the cost of the agents themselves low. We expect that agents of this type, which search efficiently over programs under limited resources, will increasingly accelerate autonomous scientific discovery~\cite{lu2026aiscientist}, since resource optimization enables faster exploration within the same budget and can thus substantially aid the discovery process.

\section{Methods}

\subsection{Problem formulation}

AIBuildAI-2.5 addresses automated AI model building as a \emph{code search} problem. Given a space of candidate programs and a reward function $R$ that scores a program by executing it, the goal is to find the program with the highest reward. For AI model building specifically, a program is usually a training-and-inference pipeline that trains a model on the training data, and the reward is the trained model's performance, such as its accuracy on a validation set. This space is large, so the efficiency and effectiveness of the search algorithm are critical. Code search of this kind has discovered programs for problems ranging from mathematics and algorithm design~\cite{FunSearch2024,Novikov2025AlphaEvolveAC} to automated machine learning~\cite{du2026mlevolve}, and among search paradigms, tree search has become one of the most widely adopted for automated AI model building. Such a search maintains a tree in which each node $n$ is a program and the root $n_0$ is an empty program or a task-provided template. Each node can be executed to obtain its reward $R(n)$, and the search grows the tree by generating a child $n'$ from a parent $n$ in an attempt to improve on $R(n)$. Tree search has two advantages: every new candidate is derived from an earlier one and thus inherits what previous attempts have learned, and separate branches explore different directions independently of one another. AIBuildAI-2.5 therefore adopts this search strategy. Within this framework, the effectiveness of the search largely depends on the design of two major operations: (1) expansion of a tree node, that is, given a parent node, how to generate a child node carrying new code; and (2) selection of a tree node to expand, that is, given the current tree with its many nodes, which node is the most promising to expand next. The two operations alternate until the budget is exhausted, after which the best program found is returned.

AI model building differs from other code-search problems in the cost of a single evaluation, namely the execution that yields the reward $R(n)$ of a node. Whereas a task such as GPU-kernel optimization can score a program in seconds by measuring its throughput, here every evaluation requires training a full model, which can take hours of wall-clock time on scarce GPUs. The number of nodes in the search tree is therefore far smaller than in a typical code tree search. The search therefore has fewer and noisier reward scores, which demands higher quality in each generated child program and greater accuracy in selecting the node to expand. Existing tree search methods rely heavily on the rewards of child nodes to guide the search direction~\cite{du2026mlevolve,jiang2025aide}, but in this setting that signal is both scarce and noisy, which calls for a search strategy different from that of typical tree search.

\subsection{Agent-based tree search}

AIBuildAI-2.5 carries out both operations with LLM-based agents, each specified by the instructions it is prompted with and the tools it may call. An agent's work is an \emph{agentic} operation: it issues many LLM calls and invokes external tools, for example to read artifacts, search the web, write files, or run code, before returning its result~\cite{yao2023react}. A single operation can thus debug a pipeline until it runs end-to-end. Following AIBuildAI~\cite{zhang2026aibuildai}, AIBuildAI-2.5 divides the work among several specialized agents. This division keeps the context of each agent short, because an agent receives only the information its own job requires, and allows each agent to be specialized to a narrow task. Both properties improve performance relative to a single agent that performs the entire workflow.

For node expansion, AIBuildAI-2.5 leverages three agents, a designer, a coder, and a reviser. The \emph{designer} is prompted to read the task description and propose several distinct, high-level modeling strategies, deliberately varied so that the search begins from different regions of the program space. Its tools allow it to inspect the dataset and search the web for relevant techniques. The \emph{coder} is prompted to turn a strategy or a revision instruction into a program. With tools to read, write, and execute files in an isolated workspace, it writes the training-and-inference program, runs short smoke tests, and debugs until the code executes end-to-end. The \emph{reviser} is prompted to read a node, its program, reward, and logs, and to emit one or more textual revision instructions, each a concrete next change such as altering an architecture component, revising the data pipeline, or fixing a specific error. With read-only access to the artifacts of the node, it decides what to try next but does not write code itself, leaving implementation to the coder. At the start of the search, before any node exists to revise, the designer is invoked once at the empty root $n_0$ to obtain $N$ initial designs, and a coder turns each into a program. These become the root's children. Expanding a node thereafter is a two-step process: the reviser reads a node $n$ together with its program, reward $R(n)$, and logs and proposes one or more revision instructions, and for each instruction a coder applies it to the program of $n$ to produce a child $n'$ carrying the revised program. Once a coder has finished implementing a child, that child is a leaf of the current search tree. From all current leaves, AIBuildAI-2.5 next selects the most promising to execute and so obtain its reward.

For node selection, AIBuildAI-2.5 leverages two agents, a judge and a selector. Formally, the search holds a \emph{pending pool} of current leaf nodes, programs implemented but not yet executed, and the task is to order the pool so that the most promising leaf is executed first. The \emph{judge} acts as a local evaluator: it assesses each leaf from several perspectives and compresses the artifacts of the node into a few values that determine search priority. Given read access to the revision instruction that produced a leaf $n$ and to the surrounding tree, it emits three integer scores on a 0--5 scale, namely the expected improvement of $n$ over its parent, how well its rationale is anchored in the parent's measured results, and the computational feasibility of executing $n$ within one slot. It calibrates each dimension against outcomes already observed on sibling and ancestor nodes rather than against the claims of the candidate itself, and assigns an all-zero score to any candidate that exhibits reward hacking behavior, such as fabricating metrics or leaking validation data. The \emph{selector} then acts globally: it ranks all leaves together from their judge values and the overall state of the search. Given the judge scores of every leaf, the live tree, and the resources that remain, it returns the leaves in the order they should be executed, looking beyond the raw scores to favor architectural diversity, discount lineages whose parents hit dead ends, suppress near-duplicates, and trade exploration against exploitation as the budget shrinks. The judge and selector thus give each leaf a strong prior estimate of its promise, drawn from the knowledge of the LLM about the likely effect of the proposed change, before any execution has taken place. Because the judge calibrates against outcomes already observed and the selector re-orders the pool whenever a node finishes, this prior is combined with the real rewards the search has measured, which is the signal that Monte Carlo-style tree search relies on alone~\cite{du2026mlevolve}. With only a small number of executions, rewards alone provide a scarce and noisy ranking signal, and the prior allows the search to rank leaves before they are executed. The leaves at the head of the ordering are executed, each yielding its reward $R(n)$ and becoming a node that the reviser and coder then expand as above.

\subsection{Auxiliary agents}

Apart from the core agents introduced above, AIBuildAI-2.5 contains three auxiliary agents that make AI model building end-to-end in the fully automated setting we target. For example, MLE-Bench~\cite{chan2025mlebench} supplies only a dataset and a natural-language task description and accepts a single submitted model. The \emph{setup} agent automatically builds the software environment in which the code runs. It runs first, is prompted to read the task and inspect the data, and, with tools to run shell commands and install packages, prepares the shared environment and runtime on which the other agents rely. The \emph{data-split} agent creates the validation set on which every program is scored. It is prompted to partition the given training data into a training split and a global validation set, with tools to read the data and write the resulting split. Fixing a single validation set shared across all nodes makes the rewards $R(n)$ comparable across nodes, as the problem formulation assumes. The \emph{aggregator} agent produces the single model that is finally submitted. It runs at termination and is prompted to combine the strongest completed nodes into one submission, by selecting the best or ensembling several~\cite{caruana2004ensemble}, with read access to those nodes' programs and predictions and tools to write the final inference artifact.

\subsection{Resource-aware job scheduler}

AIBuildAI-2.5 includes a scheduler that launches the jobs the selector has ranked while respecting the compute available, so that several nodes can potentially execute in parallel given the resource status, reducing the overall wall-clock time of the search. Given the ordered pending pool, the scheduler decides whether to launch the job at its head from the current state of the hardware. If no job is running, it launches at once. If jobs are already running, it checks for a resource bottleneck, GPU memory or utilization near capacity or CPU utilization above a threshold, and launches only when the free resources meet a predefined threshold condition, so that a new job does not deprive the running jobs of resources. After each launch, the scheduler waits for a time window before checking the resource status again, so that the newly launched job reaches a stable level of resource usage before the next decision is made.

\subsection{Model routing}

The AIBuildAI-2.5 workflow is made up of many agent calls, each serving a different purpose within a run, and any of them can be backed by a model drawn from a pool that ranges widely in cost and capability. How demanding those calls are, and therefore which model is worth its price, varies with both the task and the agent role, because the source of difficulty differs from task to task: one task hinges on implementation, a long training pipeline that must be written and debugged correctly, and benefits from the strongest coder model available, whereas another task has a routine pipeline and hinges instead on deciding what to try next, so that the designer and reviser justify a costly model while the coder does not. Serving every call with a single frontier model ignores this variation and overpays for most calls. AIBuildAI-2.5 therefore assigns each role $r$ a model $\rho(r)\in\mathcal{M}$ drawn from a pool $\mathcal{M}$ ordered from lightweight to frontier, in the spirit of recent LLM routing that trades model cost against quality~\cite{ding2024hybridllm,wang2025mixllm,zhang2025beyondgpt5,zhang2025routerr1}. Where that work routes individual queries, dispatching each call to a model chosen for it, AIBuildAI-2.5 routes at the level of the agent: the model backing an agent is fixed before the agent is activated, and it is chosen per search node, so two coder invocations in the same run may be served by different models.

The assignment is produced by a dedicated \emph{router} agent, which receives the same inputs as the agents it routes, for example the task instruction, the data directory, which its tools allow it to inspect, and a view of the search-tree record, that is, the same record that the coder reads as its lineage and that the judge and selector read in full. The router differs from a worker agent in its instruction, which asks it to judge which model each role requires rather than to solve the task. Two further inputs are specific to routing. The first is a per-run memory: a record of every node already materialized in the current run, giving the model that served each earlier agent and the score its node reached. Its purpose is to let the router adjust its choices within a run from these outcomes, for example by escalating the coder to a stronger model when an ancestor on the current lineage was coded by a cheap model and underperformed, or by diversifying away from a model that a sibling has already used. The second is the router knowledge system described below. From these inputs, the router outputs one model for every routed role, choosing the cheapest model in the pool that it judges sufficient for the task before that model is invoked~\cite{cao2026scope}.

Whereas the per-run memory carries experience within a run, the router knowledge system carries experience across runs and grows with every run the agent completes. It shares its two-level design with the evolving knowledge system of AIBuildAI-2~\cite{zhang2026aibuildai2}, and accumulates knowledge about which model should serve which role. The upper level holds distilled knowledge cards, each indexed by a model, an agent role, and a task condition, where the condition describes the property of the task to which the card applies, such as a strict submission schema or a hard image-classification problem. Each card carries a verdict on whether that model is sufficient, insufficient, or required for that role under that condition, together with references to the runs that support it. The lower level holds those runs, namely the per-task evidence records cited by the cards and the raw outcomes of each run, such as its cost, the metric it achieved, and whether its submission was valid~\cite{xu2025amem}. Every completed run writes its outcomes back to this level, so that the knowledge system keeps growing with new runs. In addition, when routing a new task, the router can leverage this routing knowledge from existing tasks with similar properties.

\subsection{Baselines}

The principal automatic AI model building agent we compare against is MLEvolve~\cite{du2026mlevolve}, which likewise searches over a tree of candidate programs. Each node is a program, and the tree is expanded by an operator that is a single LLM call: given a parent program, it samples a modified program in one generation step, whether drafting a new solution from the root, improving a promising node, or debugging a broken one. Which node to expand is chosen by an upper-confidence-bound-for-trees (UCT) rule in the spirit of Monte Carlo tree search: each node is scored by the average reward backed up from its descendants plus an exploration bonus that shrinks as the node accrues visits, so that the search favors nodes that have scored well and nodes that have been tried little. In detail, each executed node contributes a single scalar reward derived from its validation metric, with failing nodes penalized, which is propagated up the tree into the visit counts and summed rewards the rule consumes. As the run enters its later stage, the rule shifts from exploration toward exploitation and increasingly prioritizes nodes with higher summed rewards. As an additional feature, when several branches have each produced strong solutions and progress stalls, a fusion step recombines them into a new candidate. Unlike MLEvolve, AIBuildAI-2.5 conducts expansion and selection with specialized agents instead of a single LLM call and a heuristic rule, respectively: the coder debugs each program until it executes end-to-end, improving the success rate of each node, and the judge and selector add an LLM prior to the measured rewards, improving the reliability of selection when only a small number of executions are available. AIBuildAI-2.5 further schedules jobs according to the current hardware state and routes each agent role to the least costly sufficient model, lowering the cost of a run.

\subsection{Experimental settings}

For both the MLE-Bench and AIRS-Bench experiments, AIBuildAI-2.5 routes each agent role over a pool of three Claude models, namely Claude Haiku 4.5, Claude Sonnet 4.6, and Claude Opus 4.7~\cite{anthropic2026claude47}, seeds each task with $N=7$ initial drafts, and runs under a 24-hour wall-clock budget on a Linux x86-64 machine with 24 vCPUs, 256\,GB of RAM, and one NVIDIA A100 GPU. For MLE-Bench, evaluation follows the MLE-Bench protocol~\cite{chan2025mlebench} and baseline numbers are taken from the official MLE-Bench leaderboard~\cite{mlebench_commit_2026}. For AIRS-Bench, the baseline agent MLEvolve~\cite{du2026mlevolve} uses Claude Opus 4.7 as its backbone LLM and runs under a 24-hour wall-clock budget on a Linux x86-64 machine with 24 vCPUs, 256\,GB of RAM, and one NVIDIA A100 GPU. Both agents are evaluated following the official AIRS-Bench protocol, using the evaluation script shipped with each task~\cite{lupidi2026airsbench}.

\section*{Data availability}
The MLE-Bench benchmark data used in this study are publicly available from the MLE-Bench repository at \url{https://github.com/openai/mle-bench}, and the AIRS-Bench benchmark is publicly available at \url{https://github.com/facebookresearch/airs-bench}.

\section*{Code availability}
The AIBuildAI-2.5 software is publicly available at \url{https://github.com/aibuildai/AI-Build-AI/releases/tag/v2.5-latest}. Detailed instructions for using the system are provided at \url{https://github.com/aibuildai/AI-Build-AI}.

\section*{Author contributions}
P.Q., R.Z., Q.C., H.G., and L.Z. contributed to conceptualization, methodology, software, investigation, analysis, writing---original draft, and writing---review and editing. P.X. contributed to conceptualization, methodology, investigation, analysis, writing---original draft, and writing---review and editing.

\section*{References}
\begingroup
\renewcommand{\bibsection}{}
\bibliographystyle{unsrtnat}
\bibliography{ref}
\endgroup

\end{document}